\documentclass{article}

\usepackage[preprint]{neurips_2024}
\usepackage{amsmath}
\usepackage{amsmath,amssymb,amsthm}
\usepackage{graphicx}
\usepackage{microtype}
\usepackage{booktabs}
\usepackage{adjustbox}
\usepackage{multicol}
\usepackage{multirow}
\usepackage{booktabs}  
\usepackage{colortbl}  
\usepackage{amssymb}   
\usepackage{pifont}    
\usepackage{booktabs}
\usepackage{adjustbox}
\usepackage{caption}

\newcommand{\cmark}{\ding{51}} 
\newcommand{\xmark}{\ding{55}} 

\usepackage[utf8]{inputenc} 
\usepackage[T1]{fontenc}    
\usepackage{hyperref}       
\usepackage{url}            
\usepackage{booktabs}       
\usepackage{amsfonts}       
\usepackage{nicefrac}       
\usepackage{microtype}      
\usepackage{xcolor}         
\usepackage{amssymb}          
\usepackage{graphicx}         
\usepackage{adjustbox}        
\usepackage{makecell}         
\usepackage{booktabs}         
\usepackage{array}            

\usepackage{pifont}           

\usepackage[utf8]{inputenc} 
\usepackage[T1]{fontenc}    
\usepackage{hyperref}       
\usepackage{url}            
\usepackage{booktabs}       
\usepackage{amsfonts}       
\usepackage{nicefrac}       
\usepackage{microtype}      
\usepackage{xcolor}         

\title{Clinical ModernBERT: An efficient and long context encoder for biomedical text}

\author{%
  Simon A. Lee\\
  Department of Computational Medicine\\
  UCLA\\
  \texttt{simonlee711@g.ucla.edu} \\
  \And
  Anthony Wu \\
  Department of Computational Medicine\\
  UCLA\\
  \texttt{anthonytkwu@g.ucla.edu} \\
  \AND
  Jeffrey N. Chiang \\
  Department of Computational Medicine \& Neurosurgery\\
  UCLA\\
  \texttt{njchiang@g.ucla.edu} \\
}

\begin{document}

\maketitle

\begin{abstract}
We introduce Clinical ModernBERT, a transformer-based encoder pretrained on large-scale biomedical literature, clinical notes, and medical ontologies, incorporating PubMed abstracts, MIMIC-IV clinical data, and medical codes with their textual descriptions. Building on ModernBERT \citep{warner2024smarter}—the current state-of-the-art natural language text encoder featuring architectural upgrades such as rotary positional embeddings (RoPE), Flash Attention, and extended context length up to 8,192 tokens—our model adapts these innovations specifically for biomedical and clinical domains. Clinical ModernBERT excels at producing semantically rich representations tailored for long-context tasks. We validate this both by analyzing its pretrained weights and through empirical evaluation on a comprehensive suite of clinical NLP benchmarks.

\end{abstract}

\section{Introduction}

Since the introduction of BERT (Bidirectional Encoder Representations from Transformers) in 2018, encoder-only transformer architectures have remained foundational to both industry-scale and research-driven natural language processing (NLP) \citep{devlin2019bert}. Although recent advances have centered around large-scale decoder-only models such as GPT \citep{radford2018improving, achiam2023gpt}, LLaMA \citep{touvron2023llama}, and DeepSeek \citep{guo2025deepseek}, prized for their generative capabilities, BERT and its derivatives continue to play a central role in real-world applications. Its sustained popularity can be attributed to its versatility and effectiveness across critical tasks, including dense retrieval \citep{gao2023retrieval}, content moderation and classification \citep{kowsari2019text}, and the extraction of sensitive or regulated information in compliance-driven environments \citep{nadeau2007survey}.

While decoder-based models dominate applications requiring coherent generation and fluent language synthesis, encoder-only transformers offer unique advantages rooted in their bidirectional attention mechanisms. Unlike causal decoders, BERT-style models allow each token to attend to both preceding and succeeding context, yielding semantically enriched embeddings. This bidirectional encoding has proven especially valuable in scenarios that depend on fine-grained semantic discrimination. Furthermore, architectural advances in recent years have significantly modernized the encoder stack, with innovations in computational efficiency (e.g., Flash Attention \citep{dao2022flashattention}), extended sequence modeling, and parameter optimization, reaffirming the relevance of the encoder paradigm.

ModernBERT \citep{warner2024smarter} exemplifies this next-generation design, achieving a notable Pareto improvement over the original BERT across speed, memory footprint, and representational fidelity. The architecture incorporates rotary positional embeddings (RoPE) \citep{su2024roformer}, GeGLU activation functions \citep{shazeer2020glu}, bias-free linear transformations for parameter efficiency \citep{dayma2021dall, xie2023empirical}, and Flash Attention \citep{dao2022flashattention} for high-throughput inference. Critically, it supports context lengths up to 8,192 tokens, facilitating rich encoding of long-form documents. Unlike prior encoders, ModernBERT also includes source code in its training corpus, extending its utility to tasks such as code search and intelligent development assistance.

Building on these architectural improvements, we present \emph{Clinical ModernBERT}, a domain-specialized encoder pretrained on biomedical literature, clinical narratives, and structured medical ontologies. Inspired by the design philosophy of BioClinicalBERT \citep{alsentzer2019publicly} but grounded in the methodological advances of ModernBERT, our model is tailored for high-accuracy understanding in long-context biomedical and clinical NLP tasks. These include clinical information retrieval, narrative classification, and domain-specific entity and relation extraction. Clinical ModernBERT reaffirms the relevance of encoder-only architectures in the age of large generative models, offering a performant, efficient, and scalable foundation for language understanding in high-stakes medical settings.

\begin{table*}[t]
    \centering
    \caption{\textbf{Architecture differences: }Comparison of encoder architectures across domains and design innovations. Clinical ModernBERT inherits ModernBERT’s architectural advances while adapting to biomedical and clinical corpora.}
    \definecolor{highlightcol}{gray}{0.9}
    \adjustbox{max width=\textwidth}{
    \begin{tabular}{l|cc>{\columncolor{highlightcol}}c}
        \toprule
        \textbf{Feature} & \textbf{BERT} & \textbf{ModernBERT} & \textbf{Clinical ModernBERT (Ours)} \\
        \midrule
        Domain Adaptation & \xmark & \xmark & \cmark  \\
        Tokenizer & WordPiece & BPE & BPE (Additional Clinical Tokens) \\
        Sequence Length & 512 & 8192 & 8192 \\
        Positional Encoding & Sinusoidal (Learned) & Rotary (RoPE) & Rotary (RoPE) \\
        Attention Mechanism & Standard & Flash Attention & Flash Attention \\
        Activation Function & GELU & GeGLU & GeGLU \\
        Bias Parameters & Present & Removed & Removed \\
        Pre-Training Corpus & BooksCorpus, Wikipedia & Natural Language + Code & 40 million PubMed + MIMIC-IV Notes \\
        Application Focus & General NLP & General NLP + Code &  Biomedical, Clinical NLP, Medical Codes \\
        \bottomrule
    \end{tabular}
    }
    \label{tab:model_comparison}
\end{table*}

\begin{table*}[t]
\centering
\caption{\textbf{Result Reporting: }Comparison of Clinical ModernBERT against prior works across pretraining supervision, ontology integration, task coverage. Clinical ModernBERT uniquely contributes comprehensive pretraining analysis and structured code ontology integration, along with strong performance across both short and long-context clinical tasks.}
\label{tab:clinical_modernbert_comparison}
\renewcommand{\arraystretch}{1.2}
\begin{adjustbox}{max width=\textwidth}
\begin{tabular}{lcccc}
\toprule
\textbf{Model} & \makecell[c]{Pretraining MLM\\Performance Reported} & \makecell[c]{Medical Code\\Ontology Integration} & \makecell[c]{Biomedical NLP\\Tasks} & \makecell[c]{Long-Context Tasks\\(i2b2 Benchmarks)} \\
\midrule
BioClinicalBERT \citep{alsentzer2019publicly} & \xmark & \xmark & \checkmark & \checkmark  \\
BioBERT \citep{lee2020biobert} & \xmark & \xmark & \checkmark & \checkmark  \\
Clinical Longformer \citep{li2022clinical} & \xmark & \xmark & \checkmark & \checkmark \\
\textbf{Clinical ModernBERT (Ours)} & \checkmark & \checkmark & \checkmark & \checkmark \\
\bottomrule
\end{tabular}
\label{tab2}
\end{adjustbox}
\end{table*}

\textbf{Contributions:}
We introduce Clinical ModernBERT, a domain-specialized transformer encoder that integrates the architectural advancements of ModernBERT with domain-adaptive pretraining over biomedical literature, clinical notes, and structured medical ontologies. Through targeted pretraining on 13 billion tokens, Clinical ModernBERT captures both granular medical terminology and the global discourse structure of clinical documentation. A token-aware masking strategy further emphasizes semantic learning over high-value biomedical spans. A summary of model features is provided in Table \ref{tab:model_comparison} with a summary of results provided in Table \ref{tab2}. 

Empirically, we demonstrate that Clinical ModernBERT is competitive or outperforms domain baselines—including BioBERT, BioClinicalBERT, and Clinical Longformer—across a suite of downstream clinical NLP benchmarks, including semantic retrieval, classification, and entity recognition. It also achieves state-of-the-art performance on long-context tasks such as i2b2 concept extraction. Latent space visualizations confirm improved alignment with clinical ontologies, showcasing its capacity to internalize medical taxonomies. By making Clinical ModernBERT and its tokenizer publicly available, we provide a scalable and high-fidelity encoder backbone for clinical NLP and biomedical research applications.

\section{Related Works}

\subsection{Biomedical and Clinical Adaptations of BERT}
The introduction of BERT in 2018 revolutionized NLP by providing a powerful general-purpose bidirectional encoder \citep{devlin2019bert}. This milestone led to a wave of domain-adapted models, particularly in the biomedical and clinical NLP communities. Among the earliest was BioBERT, which continued BERT's pretraining on large-scale biomedical corpora, including PubMed abstracts and PMC articles \citep{lee2020biobert}. BioBERT significantly outperformed vanilla BERT on core biomedical tasks such as named entity recognition (NER), relation extraction, and question answering.

Another notable variant is SciBERT, trained on a large corpus of scientific publications spanning both computer science and biomedical domains \citep{beltagy2019scibert}. Unlike BioBERT, SciBERT introduced a new domain-specific vocabulary, further boosting performance on scientific NLP benchmarks. These models demonstrated that pretraining on domain-specific corpora yields stronger performance across a wide range of biomedical tasks, particularly when encoding specialized terminology and complex discourse structures.

In parallel to biomedical literature-focused models, there has been considerable interest in adapting BERT to clinical narrative data. ClinicalBERT was one of the first models explicitly trained on clinical notes from the MIMIC-III dataset \citep{huang2019clinicalbert}. It captured the unique language patterns found in real-world clinical documentation, such as discharge summaries.

An effective strategy that gained traction was sequential domain adaptation—starting with a biomedical model and continuing pretraining on clinical notes. BioClinicalBERT exemplifies this approach, extending BioBERT by further pretraining on MIMIC-III, thereby bridging the linguistic gap between formal biomedical writing and informal, abbreviation-heavy clinical narratives \citep{alsentzer2019publicly}. Similarly, BlueBERT combined PubMed and MIMIC-III corpora in its pretraining pipeline and was benchmarked on the BLUE evaluation suite \citep{peng2019transfer}, highlighting the benefits of cross-domain fusion in biomedical understanding.

\subsection{Ontology-Enriched and Scaling Context Length BERT Variants}

Another line of work incorporates structured medical knowledge into the pretraining objective. SapBERT leverages synonym mappings from the Unified Medical Language System (UMLS) to optimize entity embeddings via contrastive learning \citep{liu-etal-2021-self}. This enables both better within-language medical entity representations and improved cross-lingual alignment, providing value for multilingual biomedical NLP systems. Similarly DK-BEHRT \citep{an2025} found that including medical codes and their descriptions and introducing them during pre-training resulted in learning a robust latent space of ICD-9 disease codes.

Although BERT and its biomedical derivatives are limited to 512 tokens, clinical documents often exceed this threshold. To address this, Clinical Longformer was introduced with support for sequences up to 4,096 tokens \citep{li2022clinical} extending the work of the longformer \citep{beltagy2020longformer}, and bigbird \citep{zaheer2020big}. By incorporating sparse attention mechanisms \citep{tay2020sparse}, it enabled long-context processing across patient narratives, longitudinal EHR entries, and radiology reports. 

Recent work has emphasized rethinking the BERT architecture itself to enhance scalability and pretraining throughput, as exemplified by MosaicBERT \citep{portes2023mosaicbert}. Soon after, ModernBERT was introduced, as a refreshed encoder-only design that incorporates rotary positional embeddings (RoPE), GeGLU activations, Flash Attention, and support for extended context lengths of up to 8,192 tokens \citep{warner2024smarter}. Additional architectural and training details are provided in the Methods section. 

Building on this emerging paradigm of modernized BERT encoders, we integrate effective strategies from prior models to develop a long-context, compute-efficient transformer that serves as a drop-in replacement for BioClinicalBERT \citep{alsentzer2019publicly}, which has long been the de facto standard in biomedical NLP. Our model overcomes BioClinicalBERT’s 512-token limitation, and is far more efficient enabling robust encoding of longer clinical and biomedical documents at scale.

\section{Methods}

\subsection{Summarizing ModernBERT}
ModernBERT diverges from the original BERT paradigm through several innovative architectural and methodological additions. These enhancements serve as the backbone for Clinical ModernBERT and ensure improved performance and scalability. The Table \ref{tab:model_comparison} summarizes the key distinctions between BERT, ModernBERT, and our proposed Clinical ModernBERT. 

\paragraph{Rotary Positional Embeddings (RoPE)}
Traditional BERT embeddings incorporate sinusoidal positional information, where position vectors $PE(\cdot)$ are added to token embeddings. In contrast, ModernBERT uses RoPE \citep{su2024roformer}, which applies rotation to pairs of embedding dimensions. Specifically, the self-attention computation is modified by rotation matrices $R^\theta$:
\[
\text{Attention}(Q, K, V) = \text{softmax}\biggl(\frac{(Q \cdot R^\theta)\,(K \cdot R^\theta)^T}{\sqrt{d_k}}\biggr)\,V.
\]

This formulation embeds relative positional information directly into the inner product of queries and keys, preserving the order-sensitive structure of input sequences without explicit position embeddings. Unlike absolute or additive positional embeddings—which often struggle to extrapolate beyond their training context length—RoPE enables the attention mechanism to capture token relationships in a position-invariant and extrapolatable way. Mathematically, RoPE encodes position via complex rotation: each dimension pair $(x_{2i}, x_{2i+1})$ is rotated by a phase proportional to position $p$, such that the inner product $\langle Q, K \rangle$ reflects relative position differences.

This encoding preserves inner product symmetries and relative distances, which becomes crucial when dealing with long sequences where the absolute position values may lie far outside the training regime. Empirically, RoPE has been shown to allow better generalization to longer contexts and reduced degradation in attention signal at distant token pairs.

\paragraph{GeGLU Activation Layers}  
Whereas standard BERT uses ReLU-based feedforward transformations,  
\[
\text{FFN}(x) = \text{max}(0,\, x W_1 + b_1)\,W_2 + b_2,
\]  
ModernBERT uses GeGLU \citep{shazeer2020glu}:  
\[
\text{GeGLU}(x,\,W,\,V,\,b,\,c) = \bigl(\text{GeLU}(xW + b)\bigr) \circ (xV + c),
\]  
where $\circ$ denotes element-wise multiplication. This gating mechanism improves representational capacity and model stability, allowing for richer feature encodings across pre-training corpora. Compared to ReLU, GeGLU introduces a learned multiplicative interaction between nonlinear and linear projections, enabling finer control over information flow through the network. Additionally, the smooth curvature of GeLU avoids the harsh saturation behavior of ReLU, mitigating gradient sparsity and improving convergence dynamics during optimization.

\paragraph{Bias Removal and Parameter Efficiency}
Bias terms are removed throughout the ModernBERT architecture, inspired by findings in \cite{dayma2021dall}, which advocate for architectural minimalism by eliminating parameters that contribute marginally to performance. In standard transformer layers, bias terms appear in every linear transformation (e.g., $Wx + b$), but empirical studies have shown that these terms often have negligible impact when layer normalization is applied—especially in large-scale pretraining regimes.

By removing these bias terms, the model reduces the total number of trainable parameters without affecting expressive capacity. More importantly, this simplification leads to a tighter optimization landscape with fewer degrees of freedom, thereby improving gradient signal consistency and enabling faster convergence. From an efficiency standpoint, bias removal yields small but compounding reductions in memory footprint and compute cost, which become non-trivial when scaled across billions of tokens.

\newpage

\paragraph{Flash Attention}
One of the chief computational bottlenecks in self-attention is its quadratic memory and compute cost with respect to input sequence length, due to the dense $n \times n$ attention matrix formed between all token pairs. ModernBERT mitigates this via Flash Attention \citep{dao2022flashattention}, which rewrites the attention computation to be both memory-efficient and hardware-aware. Rather than materializing the full attention matrix in memory, Flash Attention computes attention scores and their softmax-normalized outputs \emph{blockwise} using a tiling scheme that fits into GPU on-chip SRAM (shared memory). Specifically, attention is computed as:
\[
\text{Attention}_{\text{blockwise}}(Q,K,V) \approx \bigoplus_{i} \text{softmax}\Bigl(\frac{Q_i K_i^T}{\sqrt{d_k}}\Bigr) V_i,
\]
where $Q_i$, $K_i$, and $V_i$ are local blocks of the query, key, and value matrices, and $\oplus$ denotes concatenation across segments.

This approach leverages two key insights. First, attention can be computed in a numerically stable, streaming fashion by fusing softmax and matrix multiplication into a single pass, eliminating the need to store intermediate attention weights. Second, by structuring computation to align with memory hierarchies (e.g., keeping blocks in registers and shared memory), Flash Attention maximizes throughput while reducing reliance on high-latency global memory. The result is near-linear scaling in memory usage with respect to sequence length, without compromising exactness of the attention result.

\paragraph{Extended Sequence Length and Diverse Data}
As a result of the optimizations described, ModernBERT extends the standard BERT context length from 512 to 8,192 tokens, drastically expanding its ability to parse and understand longer contexts which have proven to be beneficial in clinical applications \citep{wornow2024context}. These features allow the model to have advantages over previous iterations of encoder based models with longformer \citep{beltagy2020longformer} having the next largest context length 50\% shorter than modernBERT.

\subsection{Pretraining Data Sources}

To facilitate robust domain adaptation of our Clinical ModernBERT model, we curated a composite corpus spanning unstructured biomedical literature, clinical free-text, and structured medical terminologies. This multi-source dataset is designed to maximize coverage across both academic and operational aspects of biomedical language. This pretraining setup combines the data sources proposed in \citet{alsentzer2019publicly} and \citet{an2025}, integrating both clinical narratives and structured medical ontologies to enhance domain-specific language modeling.

\paragraph{PubMed Abstracts}

We leveraged approximately 40 million PubMed abstracts, encompassing biomedical publications through 2025 \citep{lu2011pubmed}. These abstracts reflect the linguistic and conceptual heterogeneity of peer-reviewed biomedical literature, providing a high-coverage substrate for encoding domain-relevant semantics. The inclusion of this corpus enables the model to internalize terminology, syntax, and semantic associations prevalent in formal scientific writing across biomedical subjects. Models like sci-bert have found that including scientific literature helps learn domain-specific concepts very well \citep{beltagy2019scibert}.

\paragraph{MIMIC-IV Clinical Notes}

We incorporated deidentified clinical notes from the MIMIC-IV dataset \citep{johnson2023mimic}, drawn from real-world inpatient encounters at the Beth Israel Deaconess Medical Center. Our pretraining corpus includes discharge summaries and radiology reports. Discharge summaries encode high-density longitudinal narratives covering diagnostic reasoning, therapeutic trajectories, and patient-specific contextualization. Radiology reports introduce a specialized modality of medical interpretation characterized by compressed syntax, diagnostic speculation, and anatomical specificity.

\paragraph{Structured Medical Ontologies}

\begin{figure}[t!]
    \centering
    \includegraphics[width=\linewidth]{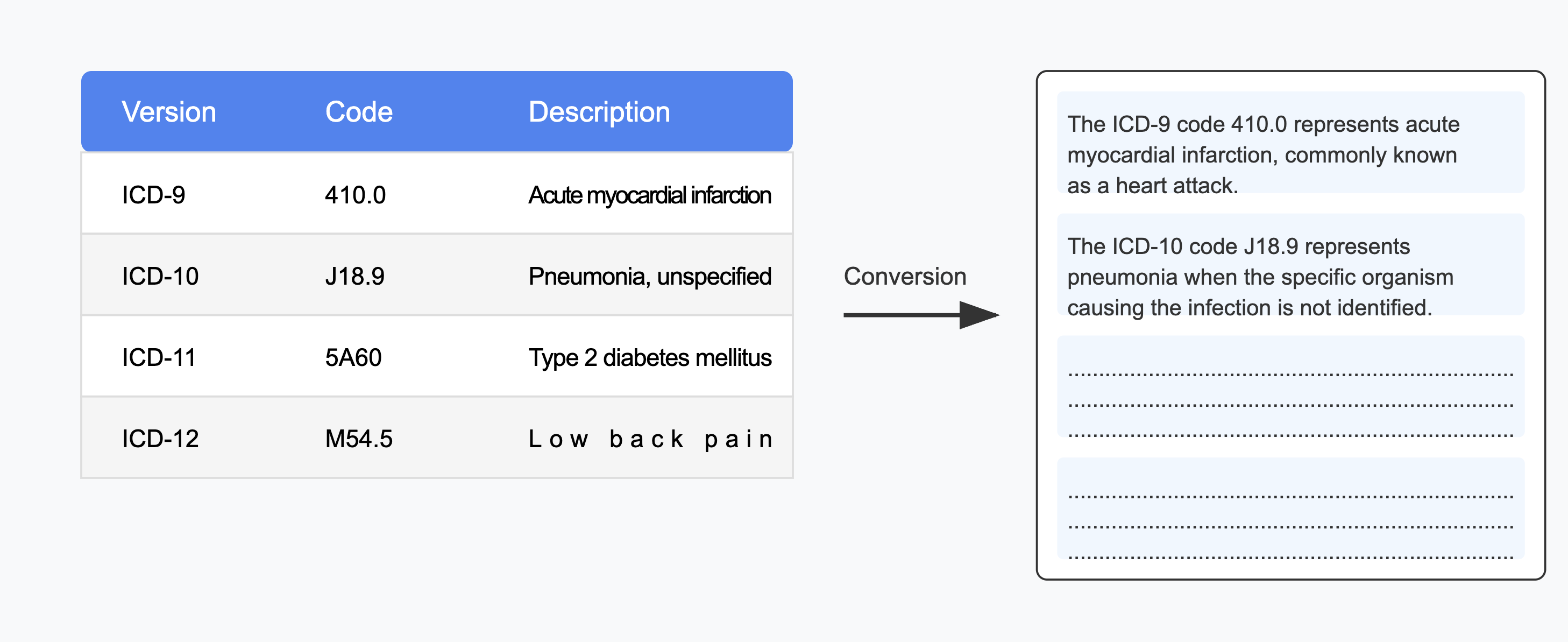}
    \caption{\textbf{Medical Code Ontologies Construction:} An illustration of structured ontology construction across multiple ICD code versions. Each row represents a distinct medical concept identified by its version-specific code and description, which is then converted into a standardized, descriptive natural language representation. This process facilitates alignment and interoperability across evolving coding schemes. This setup is inspired by methods like \citep{hegselmann2023tabllm, ono2024text} which use text templates to serialize tabular data.}
    \label{icd2}
\end{figure}

To complement the free-text sources, we integrated a comprehensive collection of standardized medical codes and descriptions, encompassing multiple revisions of the ICD taxonomy (ICD-9 through ICD-12), medication codes, and procedural terminologies (e.g. CPT). This is in light of both realizing and learning that coded language can be learned by providing natural language descriptions in conjunction with the code \citep{an2025, lee2024emergency}. This process is demonstrated in Figure \ref{icd2}. Each code-description pair is treated as a linguistic signal representing formalized medical knowledge. This structured input scaffolds the model’s understanding of discrete clinical concepts, reinforces terminology normalization, and enhances interoperability with evolving coding systems. These ontological alignments are essential for downstream tasks involving entity normalization, coding prediction, and concept linking \citep{lee2024can, soroush2024large}.

\subsection{Pre-training Optimization}

Clinical ModernBERT was pre-trained using a masked language modeling (MLM) framework designed to enhance its capacity to encode biomedical and clinical language. Initialization from the ModernBERT-base checkpoint conferred architectural advantages, including Flash Attention, rotary positional embeddings (RoPE), and GeGLU activation layers. Building on the approach of \citet{alsentzer2019publicly}, the goal was to enrich the model with specialized domain knowledge while preserving strong contextual reasoning capabilities.

The training corpus was constructed by aggregating (i) discharge summaries and radiology reports from MIMIC-IV \citep{johnson2023mimic}, (ii) approximately 40 million PubMed abstracts published through 2025, and (iii) structured medical ontologies such as ICD and CPT codes. After normalization and preprocessing, the final dataset encompassed 13 billion tokens. To prioritize semantically dense content, samples with coherent discourse—such as detailed clinical narratives—were upsampled, providing a richer training signal. This composition enabled the model to span a broad spectrum of clinical language, from fine-grained medical terminology to higher-level documentation across radiology and discharge summaries. Tokenization used a byte-pair encoding (BPE) scheme initialized from ModernBERT-base. 

To reinforce learning of clinically relevant semantics, a custom data collator was introduced that applied token-aware masking to biomedical entities during training. This collator implemented a dynamic corruption schedule, linearly decaying the masking rate from 30\% to 15\% over training. Early phases emphasized challenging and informative terms—such as medication names, procedure codes, and morphological descriptors—while later stages aimed to stabilize representation learning. This targeted perturbation strategy encouraged the model to develop nuanced contextual embeddings grounded in clinical discourse.

Training was conducted for 150,000 steps. While initial plans considered step counts derived from dataset size and average sample length, empirical evaluation showed MLM performance plateauing beyond this point. Checkpoints were saved every 10,000 steps to capture model evolution prior to any degradation. Consistent with observations from \citet{alsentzer2019publicly}, we found that convergence occurred relatively early in training (Appendix Figure \ref{wandb}).Druing pre-training we measured top-1, 5, 10, and 25 MLM accuracies as well as tracked loss to find the optimal model checkpoint.

\paragraph{Training Procedure} The pre-training schedule began with a sequence length of 128 tokens, leveraging large batch sizes and elevated learning rates to efficiently model short-range dependencies. StableAdamW \citep{wortsman2023stablelowprecisiontraininglargescale} was used for gradient clipping and stabilization, proving effective for large-batch optimization on GPU clusters. A cosine learning rate schedule guided dynamic adjustment of learning rates, while mixed-precision training improved memory utilization and throughput. Checkpoints were saved regularly to enable inspection and reuse of intermediate representations.

Training was distributed across NVIDIA A100 GPUs using multi-node orchestration. The final model weights and tokenizer artifacts were saved for reproducibility and future adaptation on huggingface \citep{wolf2020huggingface, simon_lee_2025} \footnote{https://huggingface.co/Simonlee711/Clinical\_ModernBERT}. Clinical ModernBERT is publicly released to support the broader community in downstream fine-tuning, domain-specific pretraining, and exploration of new clinical NLP benchmarks.

\section{Experimental Setup}

\begin{table*}[t]
\centering
\caption{\textbf{Dataset Statistics:} Statistics for downstream NLP tasks spanning both short and long context settings. Short context tasks typically fall within the standard input limits of models like BERT (512 tokens), whereas long context tasks significantly exceed this threshold, necessitating architectures capable of extended sequence modeling.
}
\begin{adjustbox}{max width=0.97\textwidth}
\begin{tabular}{l l l r r r}
\toprule
\multicolumn{6}{l}{\textbf{Short Context Tasks}} \\
\midrule
\textbf{Dataset} & \textbf{Task} & \textbf{Data Source} & \textbf{Sample Size} & \textbf{Avg. Seq. Length} & \textbf{Max Seq. Length} \\
\midrule
EHR-Prediction & Classification & MIMIC-IV ED & 400,019 & 278.6 & 2,684.0 \\
MedNER & NER & Custom & 3,655 & 17.7 & 125.0 \\
Pubmed-NCT & Multiclass Classif. & PubMed & 221,186 & 26.2 & 260.0 \\
PMC-Retrieval & Retrieval & PMC & 167,034 & 37.3 & 2,728.0 \\
\bottomrule
\end{tabular}
\end{adjustbox}
\label{data-stats}
\end{table*}

\begin{table*}[t]
\centering
\begin{adjustbox}{max width=0.83\textwidth}
\begin{tabular}{l l l r r r}
\toprule
\multicolumn{6}{l}{\textbf{Long Context Tasks}} \\
\midrule
\textbf{Dataset} & \textbf{Task} & \textbf{Data Source} & \textbf{Sample Size} & \textbf{Avg. Seq. Length} & \textbf{Max Seq. Length} \\
\midrule
i2b2 2006 & NER & i2b2 & 66,034 & 867.0 & 3,986.0 \\
i2b2 2010 & NER & i2b2 & 43,947 & 1,459.3 & 6,052.0 \\
i2b2 2012 & NER & i2b2 & 13,108 & 793.6 & 2,900.0 \\
i2b2 2014 & NER & i2b2 & 83,466 & 5,133.5 & 14,370.0 \\
\bottomrule
\end{tabular}
\end{adjustbox}
\end{table*}

\subsection{Assessing Pre-training}

Masked Language Modeling (MLM) is a self-supervised objective in which random tokens within an input sequence are replaced with a special [MASK] token, and the model is trained to reconstruct the original tokens based solely on the surrounding context. Given a sequence \( x = (x_1, \ldots, x_n) \), a subset of positions \( \mathcal{M} \subset \{1, \ldots, n\} \) is selected for masking. The model receives \( \tilde{x} \), where \( \tilde{x}_i = \text{[MASK]} \) for \( i \in \mathcal{M} \), and is trained to minimize the cross-entropy loss:

\[
\mathcal{L}_{\text{MLM}} = - \sum_{i \in \mathcal{M}} \log p_\theta(x_i \mid \tilde{x}),
\]

where \( p_\theta \) is the model's predicted distribution over the vocabulary. This objective encourages the model to capture syntactic and semantic dependencies in the data.

To quantify performance, we compute top-$k$ accuracy over masked positions, where a prediction is considered correct if the ground truth token appears in the top-$k$ most probable tokens predicted by the model. Specifically, for each masked token \( x_i \), we sort the predicted distribution and check whether \( x_i \) lies in the top $k$ logits. These metrics provide a granular view of the model’s ability to recover meaningful clinical vocabulary under varying levels of strictness, with top-1 reflecting precision and top-25 capturing broader lexical recall.

\subsection{Benchmarking Models}

We compare Clinical ModernBERT against a range of baseline and state-of-the-art models to ensure broad coverage across both context regimes. This includes BERT-base \citep{devlin2019bert} as a general-purpose baseline, BioBERT \citep{lee2020biobert} and BioClinicalBERT \citep{alsentzer2019publicly} as domain-adapted encoders, and Clinical Longformer \citep{li2022clinical} and ModernBERT \citep{warner2024smarter}, which extend BERT to support longer contexts. These baselines collectively allow us to assess the benefits of domain-specific adaptation, architectural modernization, and sequence length scaling in clinical NLP.

\subsection{Benchmarking Datasets}

To comprehensively evaluate Clinical ModernBERT, we design experiments spanning both standard biomedical NLP benchmarks and long-context clinical tasks. The former comprise widely-used short-context datasets for clinical note classification, biomedical named entity recognition (NER), and scientific trial labeling, where input sequences remain well within the 512-token constraint. These benchmarks—originally introduced in prior studies such as BioBERT \citep{lee2020biobert} and BioClinicalBERT \citep{alsentzer2019publicly}—primarily assess local representational fidelity, focusing on tasks where short-range dependencies dominate. We additionally include a sentence-level biomedical retrieval benchmark \citep{zhao2023large}, which serves as a further probe of the model’s embedding quality.

To complement this, we evaluate on long-context benchmarks designed to test Clinical ModernBERT’s ability to reason over extended sequences, a core architectural advantage of the model. These tasks include full-document NER and long-range semantic retrieval, sourced from the i2b2 shared task suite and other publicly available clinical corpora, following prior protocols from \citet{li2022clinical} and \citet{zhao2023large}. This dual benchmark strategy allows us to disentangle the contributions of local versus global context modeling, and to isolate the gains attributable to architectural extensions such as Flash Attention and rotary position embeddings. Summary statistics for each dataset, including sample size and token length distributions, are provided in Table~\ref{data-stats}. Detailed dataset descriptions appear in Appendix~\ref{data}.

\subsection{Medical Codes Visualization Protocol}

To probe the representational impact of incorporating structured medical codes during pretraining, we constructed an embedding visualization pipeline centered on ICD-9 diagnosis codes. Since both the codes and their textual descriptions were included in the pretraining corpus, this analysis serves as a targeted lens into how effectively Clinical ModernBERT captures the semantics of coded clinical language. To enable qualitative analysis of the resulting latent spaces, we projected the high-dimensional embeddings into two dimensions using t-SNE. This visualization allows us to assess whether models pretrained on structured ontological data exhibit stronger alignment with the clinical taxonomy—reflected, for instance, in the degree to which embeddings cluster by ICD chapter or semantic relatedness. By comparing Clinical ModernBERT against a general-domain baseline (ModernBERT), this setup isolates the contribution of ontology-aware pretraining to the organization and separability of medical code representations.

\subsection{Model Efficiency Benchmarking Protocol}

To assess runtime efficiency under increasing computational load, we measured the forward pass latency of Distil-BERT, BioClinicalBERT, and Clinical ModernBERT across input batches of varying sizes. Synthetic clinical text inputs were generated and standardized to a fixed sequence length of 512 tokens to eliminate confounding variability due to input length. Each model was executed in inference mode using PyTorch with automatic mixed precision (AMP) enabled. Benchmarks were conducted on a single NVIDIA A100 GPU, and total wall-clock time was measured from input tokenization through to the final hidden state output, excluding any disk or network I/O. Each experiment was repeated three times to ensure stability, and the reported timing reflects the mean runtime across runs. This benchmarking framework was designed to isolate transformer-level inference cost, enabling a direct comparison of architectural efficiency between the baseline and our proposed model.

\section{Results}

\subsection{Effectiveness of Pre-training}

\begin{table}[h!]
\centering
\caption{\textbf{MLM Top-\emph{k} accuracies: } Masked language modeling (MLM) top-\textit{k} accuracies after pretraining on clinical and biomedical corpora.}
\begin{adjustbox}{width=\linewidth}
\begin{tabular}{lcccc}
\toprule
\textbf{Metric} & \textbf{Top-1 Accuracy} & \textbf{Top-5 Accuracy} & \textbf{Top-10 Accuracy} & \textbf{Top-25 Accuracy} \\
\midrule
\textbf{Value (\%)} & 63.31 & 79.67 & 83.33 & 88.10 \\
\bottomrule
\end{tabular}
\end{adjustbox}
\label{mlm_accuracies}
\end{table}

\paragraph{Qualitative Analysis.} The top-\textit{k} accuracy metrics reflected in Table \ref{mlm_accuracies} indicate robust convergence and high recall over clinical token prediction, especially at larger \textit{k}. The top-1 accuracy of 63.31\% demonstrates strong discriminative capacity, even under high entropy masking. Moreover, the sharp increases in top-5 through top-25 accuracies—approaching 88.10\%—suggest that the model consistently ranks clinically appropriate tokens among its top candidates. This behavior is indicative of successful semantic alignment with domain-specific medical terminology. Full Wandb metrics can be seen in the Appendix Section \ref{mlm}. The concurrent decline in MLM loss over training steps and the plateauing of accuracy metrics further support the conclusion that the model learns stable, high-fidelity representations of biomedical and clinical language (Figure \ref{wandb}).

\paragraph{Ablation} 

\begin{table}[h!]
\centering
\caption{\textbf{MLM Top-\emph{k} accuracies: } Masked language modeling (MLM) top-\textit{k} accuracies after pretraining on clinical and biomedical corpora under various architectural and training ablations.}
\begin{adjustbox}{width=\linewidth}
\begin{tabular}{lcccc}
\toprule
\textbf{Configuration} & \textbf{Top-1 Accuracy} & \textbf{Top-5 Accuracy} & \textbf{Top-10 Accuracy} & \textbf{Top-25 Accuracy} \\
\midrule
\textbf{Baseline (Ours)} & 63.31 & 79.67 & 83.33 & 88.10 \\
\textbf{w/o token-aware masking} & 48.84 & 53.01 & 56.10 & 58.79 \\
\textbf{w/ 15\% Masking (over 30\%)} & 58.22 & 73.87 & 76.90 & 80.57 \\
\bottomrule
\end{tabular}
\end{adjustbox}
\label{tab:mlm_accuracies}
\end{table}

We conduct a series of ablations to isolate the impact of masking strategies on the effectiveness of Clinical ModernBERT’s pretraining. Removing token-aware masking—a strategy that prioritizes clinically salient tokens for masking—results in a steep decline across all top-$k$ accuracy metrics. The top-1 accuracy drops from 63.31\% to 48.84\%, and even top-25 accuracy falls below 59\%, indicating that naïvely masking uniformly degrades the model’s ability to predict meaningful biomedical content. This highlights the importance of directing the model’s learning signal toward domain-relevant vocabulary.

Separately, we reduce the masking ratio from 30\% to 15\%, mirroring the original BERT setup. While this results in only a moderate decline, it still substantially underperforms the baseline, with top-1 accuracy falling to 58.22\%. This suggests that higher masking rates, when coupled with token-aware selection, enhance the density of the supervision signal and encourage the model to learn more informative contextual representations. Taken together, these results affirm that both what is masked and how much is masked matter critically in domain-specific pretraining regimes.

\subsection{Benchmark Across Multiple Standard Biomedical NLP Tasks}

\begin{table*}[t!]
\centering
\caption{\textbf{Performance Across Short Context Clinical and Biomedical NLP Benchmarks.} Benchmark results for EHR Classification, PubMed-NCT, and MedNER. We report AUROC for EHR Classification and Accuracy and F1 Score for PubMed-NCT and MedNER following metrics used in prior studies.}
\begin{adjustbox}{max width=\textwidth}
\begin{tabular}{lcccccc}
\toprule
\multirow{2}{*}{Model} & \multicolumn{1}{c}{\textbf{EHR Classification}} & \multicolumn{2}{c}{\textbf{PubMed-NCT}} & \multicolumn{2}{c}{\textbf{MedNER}} \\
& AUROC & Acc & F1 & Acc & F1 \\
\midrule
bert-base-uncased        & 0.9503 & 0.8754 & \underline{0.8706} & 0.842 & 0.691 \\
biobert                  & \underline{0.9680} & \underline{0.9179} & \textbf{0.9187} & \textbf{0.909} & \textbf{0.794} \\
bioclinicalbert          & 0.9678 & 0.9145 & 0.8285 & \underline{0.849} & 0.710 \\
clinical longformer      & 0.9640 & 0.8950 & 0.8250 & 0.820 & 0.720 \\
modernbert               & 0.9677 & 0.9104 & 0.7602 & 0.695 & 0.517 \\
Clinical modernbert      & \textbf{0.9769} & \textbf{0.9209} & 0.8654 & 0.829 & \underline{0.766} \\
\bottomrule
\end{tabular}
\end{adjustbox}

\label{tab:classification_results}
\end{table*}


\begin{table}[t]
\centering
\caption{\textbf{Retrieval Performance on PMC-Patients Benchmark.} PMC-Patients Retrieval performance. We report NDCG@10, Precision@10, Recall@10, and Mean Average Precision (MAP) over 128 patient queries.}

\begin{adjustbox}{max width=0.8\columnwidth}
\begin{tabular}{lcccc}
\toprule
Model & NDCG@10 & Precision@10 & Recall@10 & MAP \\
\midrule
bert-base-uncased        & 0.0600 & 0.0079 & 0.0604 & 0.0589 \\
biobert                  & \underline{0.1956} & \underline{0.0489} & 0.1866 & 0.1827 \\
bioclinicalbert          & 0.1512 & 0.0291 & 0.1466 & 0.1441 \\
clinical longformer      & 0.1700 & 0.0420 & 0.2100 & 0.1600 \\
modernbert               & 0.1865 & 0.0463 & \underline{0.2256} & \underline{0.1839} \\
Clinical modernbert      & \textbf{0.2167} & \textbf{0.0552} & \textbf{0.2791} & \textbf{0.1982} \\
\bottomrule
\end{tabular}
\end{adjustbox}
\label{tab:pmc_retrieval_full}
\end{table}

Across all tasks, models pretrained on biomedical or clinical corpora consistently outperform general-domain baselines. On EHR classification, clinical modernbert achieves the highest AUROC (0.9769), outperforming both biobert and bioclinicalbert, and demonstrating strong generalization to structured EHR narratives. On the PubMed-NCT classification task, biobert yields the best F1 score (0.9187), though clinical modernbert surpasses it in accuracy, suggesting improved calibration. For MedNER, biobert again leads in F1 (0.794), while clinical modernbert remains competitive (0.766), outperforming all other models except biobert.

In the retrieval setting (PMC Patients), clinical modernbert achieves the best performance across all metrics, including NDCG@10 (0.2167) and MAP (0.1982), indicating that clinically informed pretraining confers advantages for semantic matching. Overall, clinical modernbert exhibits strong generalization across diverse biomedical NLP tasks, spanning classification, named entity recognition, and retrieval.

\subsection{Long Context Biomedical NLP Tasks}

On long-context clinical NER benchmarks, performance generally tracks the model's ability to process extended sequences while preserving token-level precision. Clinical longformer achieves the highest F1 on i2b2 2006 (0.973), 2010 (0.886), and 2014 (0.960), demonstrating that extending the context length optimized for long sequences yield measurable gains when context exceeds standard transformer limits. Clinical modernbert, however, shows consistently competitive performance across all datasets, achieving the best result on i2b2 2012 (0.804) and the second-best on all other benchmarks. 

In contrast, biobert and bioclinicalbert perform strongly on shorter variants of the i2b2 tasks (e.g., 2006 and 2010), but show diminishing returns as document length and entity density increase. Modernbert outperforms these biomedical baselines on all four datasets, suggesting that long-range architectural modifications, prove better than domain tuning outperfomring modes like biobert and bioclinial bert on these long-context baselines. Overall, these results indicate that both domain-specific pretraining and architectural adaptation are necessary for robust performance on long-context clinical NER tasks.

\begin{table}[t!]
\centering
\caption{\textbf{Performance on i2b2 Long Context Benchmarks.} F1 scores for different models across i2b2 benchmark datasets. Bold indicates the best performance per column, underline indicates the second-best.}

\begin{adjustbox}{max width=\textwidth}
\begin{tabular}{lcccc}
\toprule
Model (all F1) & i2b2 (2006) & i2b2 (2010) & i2b2 (2012) & i2b2 (2014) \\ \midrule
bert-base-uncased        & 0.938 & 0.834 & 0.758 & 0.927 \\
biobert                  & 0.947 & 0.866 & 0.791 & 0.929 \\
bioclinicalbert          & 0.950 & 0.860 & 0.772 & 0.928 \\
clinical longformer      & \textbf{0.973} & \textbf{0.886} & \underline{0.799} & \underline{0.960} \\
modernbert               & 0.957 & 0.875 & 0.782 & 0.948 \\
Clinical modernbert      & \underline{0.965} & \underline{0.883} & \textbf{0.804} & \textbf{0.966} \\
\bottomrule
\end{tabular}
\end{adjustbox}
\label{tab:results}
\end{table}

\subsection{Latent Space Visualizations for medical codes}

\begin{figure}[t!]
    \centering
    \includegraphics[width=\linewidth]{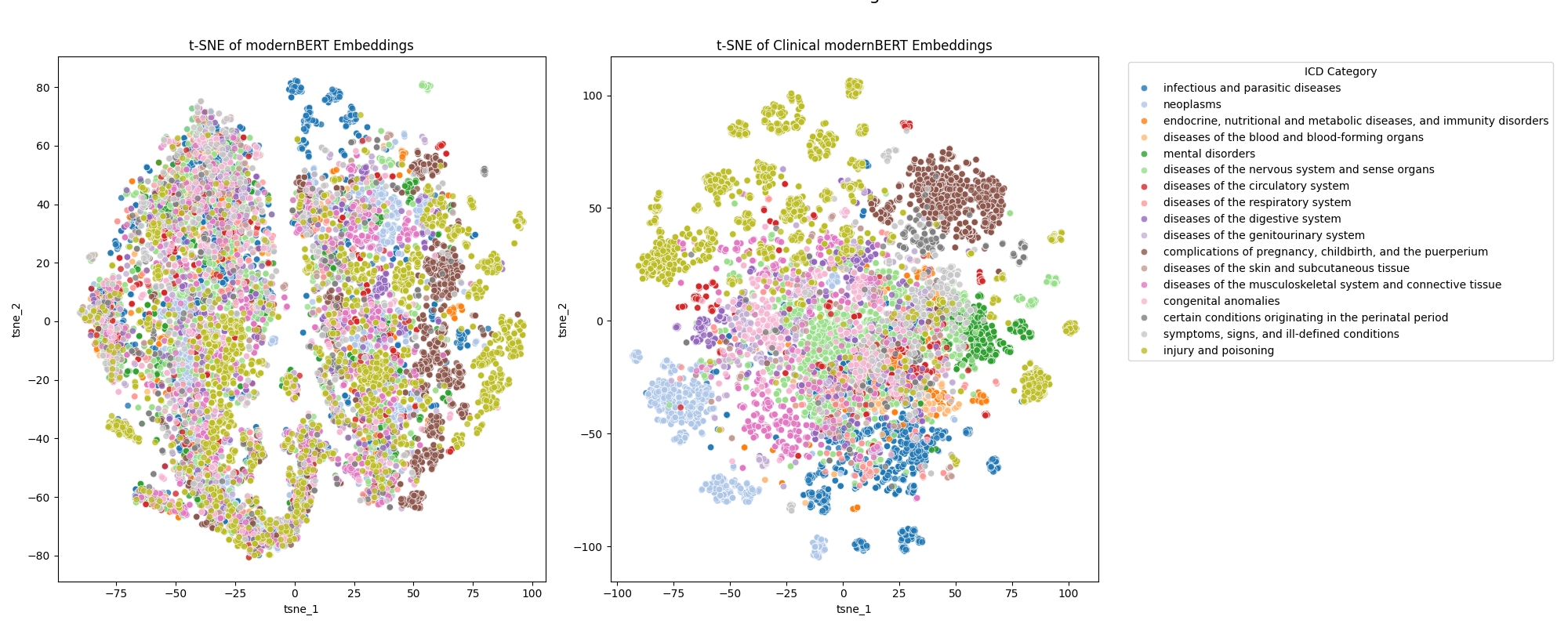}
    \caption{\textbf{ICD-9 tSNE Latent Space Visualization:} A tSNE visualization of the ICD 9 Diagnoses codes using modernBERT versus Clinical ModernBERT. This visualization provides the added use of adding the medical code ontologies as a pre-training source to encode coded language seen frequently in clinical practice.}
    \label{icd}
\end{figure}

To qualitatively assess the clinical semantic structure captured by our proposed model, we visualize t-SNE projections of diagnosis code representations derived from ModernBERT (left) and our Clinical modernBERT (right) (Figure \ref{icd}). Each point represents an ICD code embedding, color-coded by its corresponding high-level ICD category. Embeddings are extracted by feeding tokenized ICD descriptions through each model and taking the classification token (\texttt{[CLS]}) embeddings. We then project these high-dimensional embeddings into 2D using t-SNE, preserving local similarity structure.

The comparison reveals stark differences. ModernBERT, pretrained primarily on general domain corpora, fails to form well-defined clusters among diagnostic categories, particularly for ontologically proximate conditions such as respiratory and circulatory disorders. In contrast, Clinical modernBERT—augmented with structured medical code ontologies during pretraining—produces significantly cleaner separations and tighter intra-category clustering by their ICD chapters. Notably, disease categories such as neoplasms, nervous system disorders, and congenital anomalies are distinctly separated, reflecting improved semantic alignment with clinical taxonomies.

\subsection{Efficiency and Scalability of Encoding Embeddings}

\begin{figure}[t!]
    \centering
    \includegraphics[width=\linewidth]{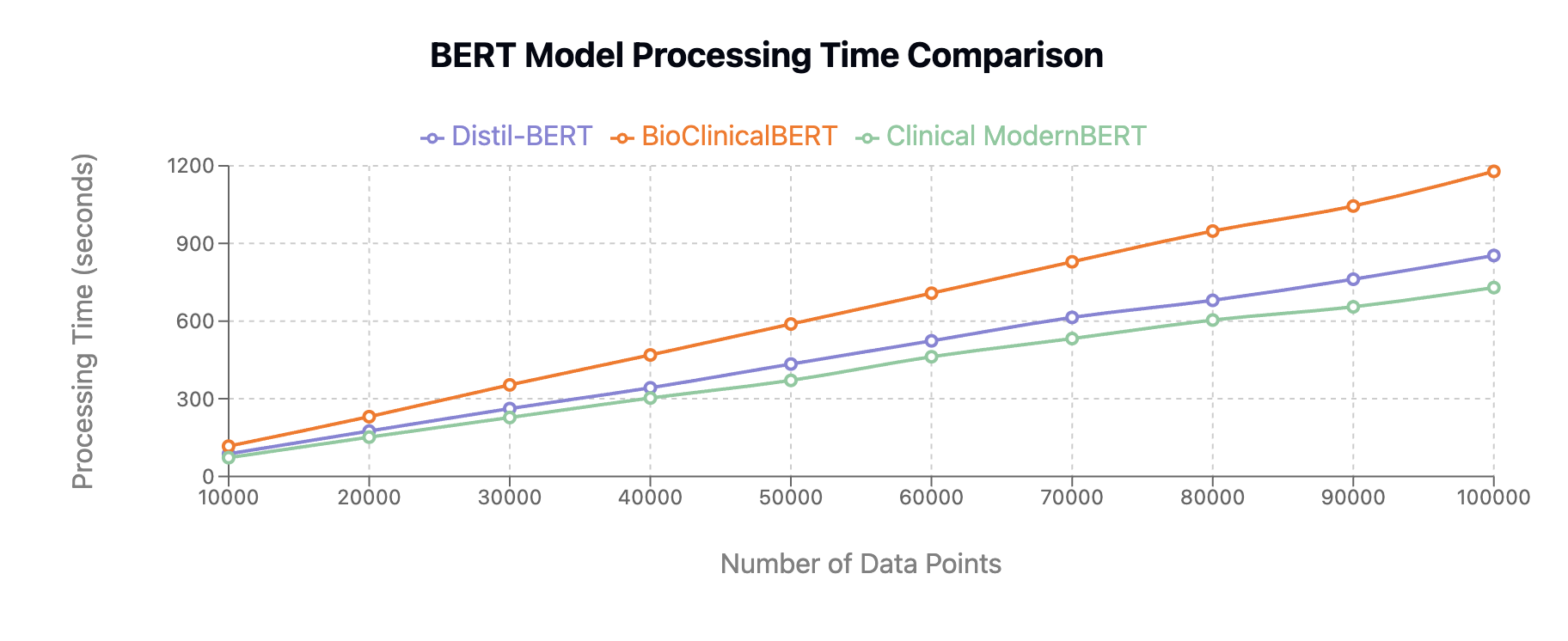}
    \caption{\textbf{Comparative Performance Analysis of BERT Models:} 
This figure demonstrates the processing time requirements across three BERT variants (Distil-BERT, BioClinicalBERT, and Clinical ModernBERT) as data volume increases from 10,000 to 100,000 points. BioClinicalBERT consistently shows the highest computational demand, requiring approximately 1.4x the processing time of Distil-BERT and 1.6x that of Clinical ModernBERT at maximum load. Clinical ModernBERT demonstrates superior efficiency, maintaining the lowest processing times across all data volumes, making it optimal for resource-constrained environments.}
    \label{time}
\end{figure}

Figure~\ref{time} presents the comparative processing times of Distil-BERT, BioClinicalBERT, and Clinical ModernBERT across increasing data volumes. Clinical ModernBERT demonstrates a clear computational advantage, maintaining the lowest runtime footprint throughout, even at scales of 100,000 data points. In contrast, BioClinicalBERT exhibits the steepest growth in processing time, with a final runtime nearly 60\% higher than Clinical ModernBERT. While Distil-BERT, by design, remains lightweight, it still trails Clinical ModernBERT in efficiency, particularly at higher loads. This suggests that architectural optimizations in Clinical ModernBERT—such as the integration of Flash Attention and linear-time positional encoding—confer measurable speedups without sacrificing model capacity. These results underscore Clinical ModernBERT's suitability for scalable deployment in clinical pipelines where latency and throughput are critical and are consistent with previous findings in the literature \citep{yamagishi2025modernbert}.

\section{Discussion}

\subsection{Domain Adaptation and Representational Quality}

The empirical findings across masked language modeling, downstream benchmarks, and latent space analyses offer converging evidence that Clinical ModernBERT achieves superior domain adaptation in the clinical and biomedical space. The elevated top-1 MLM accuracy of 63.31\%, and especially the marked increase in top-5 to top-25 accuracies, highlights the model’s capacity not only to disambiguate masked tokens precisely but also to consistently narrow in on semantically proximate candidates. This predictive precision suggests a strong inductive bias for clinical syntax and terminology, likely a result of both the corpus composition and our token-aware masking regime.

Clinical ModernBERT’s strong showing in EHR classification and PMC retrieval further validates this pretraining protocol. The model attains state-of-the-art AUROC in EHR classification and consistently outperforms baselines on all retrieval metrics. This duality—high discriminative capacity in structured tasks and nuanced semantic alignment in open-ended retrieval—indicates that the learned representations are both fine-grained and semantically rich. We also saw considerable performance gains on long-context tasks on the i2b2 datasets (2006, 2010, 2012, and 2014) where clinical longformer and clinical modernBERT had switched off between best and second-best models in that benchmark.

\subsection{Ablation Findings and Pretraining Design}

The ablation studies underscore the critical role of pretraining design decisions. Removing token-aware masking led to a catastrophic drop in performance across all top-$k$ thresholds, indicating that uniformly masking tokens fails to guide the model towards domain-salient lexical patterns. Similarly, reverting to a 15\% masking ratio degrades model efficacy, reinforcing the intuition that denser supervision—when targeted effectively—facilitates richer context modeling. These findings contribute to a growing body of evidence suggesting that in specialized domains, indiscriminate adoption of generic pretraining heuristics can be suboptimal. Instead, task and domain-specific curriculum construction plays a determinative role in representational quality.

\subsection{Latent Structure and Qualitative Insights}

The latent space visualizations provide compelling qualitative evidence that our pretraining approach captures clinically meaningful structure. The improved clustering by ICD category in Clinical ModernBERT suggests that the model internalizes not just surface-level co-occurrence statistics but also latent ontological relationships. This aligns with theoretical intuitions from distributional semantics, where models trained on structured knowledge tend to reflect the graph topology of their source corpora in their embedding geometries. On the efficiency side, we also saw that modernBERT has the ability to generate dense embeddings at a scalable efficiency outperforming BioClinical BERT as well as distil-BERT which is a smaller and faster version of the original BERT architecture reinforcing the modifciations made by \cite{portes2023mosaicbert}.

\subsection{Future Directions}

Looking ahead, we envision several promising directions. One avenue is to investigate scaling laws within the clinical pretraining regime—examining whether the predictable log-linear improvement observed in general language models extends to specialized corpora or if new inflection points emerge. We also foresee opportunities to identify the limitations of these models through a comprehensive error analysis, as proposed in previous studies \citep{lee2024enhancing, soroush2024large}. Moreover, integrating these approaches with clinical benchmarking frameworks such as MEDS, which facilitates the transformation and summarization of tabular data using the MEDS schema \citep{arnrich2024medical, kolo2024meds}, represents another compelling direction. Finally, extending Clinical ModernBERT to multimodal settings—linking textual information with imaging or waveform data—could unlock novel capabilities in clinical decision support. 

In summary, Clinical ModernBERT demonstrates that thoughtful adaptation of general language modeling principles to the clinical domain—through targeted masking, structured knowledge integration, and dense supervision—yields models that are not only competitive but, in many cases, state-of-the-art across a wide range of biomedical NLP tasks.

\bibliographystyle{plainnat}
\bibliography{references}
\newpage
\appendix
\section{Dataset Details}
\label{data}

\paragraph{Biomedical NLP tasks (short-context)}

\paragraph{EHR Prediction: Text Classification}  
To evaluate representation quality on structured clinical data, we include an emergency department (ED) disposition prediction task based on the dataset introduced in \citet{lee2024multimodal}. The dataset comprises structured EHR records from emergency visits, including demographic variables, triage vitals, chief complaints, procedures, labs, and medication events recorded during the first hour of a patient’s ED stay. These features are transformed into textual “pseudo-notes,” which emulate the style and structure of clinical documentation, making the data compatible with language models. The prediction target is the patient’s disposition outcome—i.e., whether the patient is admitted or discharged. This task serves as a proxy for real-time decision support and reflects a high-stakes, operationally critical setting in clinical care. 

\paragraph{PubMed-200k-RCT: Text Classification}  
The PubMed-200k-RCT dataset is a sentence-level classification benchmark derived from structured text in the PubMed corpus \citep{dernoncourt2017pubmed200krctdataset}. It contains approximately 200,000 sentences, each labeled according to its rhetorical function within a scientific abstract: \textit{Background}, \textit{Objective}, \textit{Methods}, \textit{Results}, or \textit{Conclusions}. The dataset is formatted with fields for abstract ID, sentence ID, label, and sentence text. 

\paragraph{Medical Entity Recongition: NER: Named Entity Recognition at token level}  
The MedNER dataset provides a supervised benchmark for named entity recognition (NER) in clinical and biomedical text \footnote{https://www.kaggle.com/datasets/arunagirirajan/medical-entity-recognition-ner}. We frame this dataset as a resource for token-level medical entity recognition under constrained supervision. The MedNER corpus comprises tokenized clinical and biomedical text annotated for binary entity presence, where each token is labeled as either part of a named medical entity or not. This reductionist framing positions the dataset as a testbed for probing entity boundary detection and semantic salience in the absence of hierarchical supervision. It is particularly well-suited for studying low-resource NER, binary token classification, and early-stage entity bootstrapping in noisy medical corpora, and can serve as a diagnostic task within broader pretraining or adaptation pipelines.

\paragraph{PMC-Patients: Retrieval}  
The PMC-Patients dataset provides a large-scale benchmark for retrieval-based clinical decision support tasks, leveraging real-world case reports from PubMed Central (PMC) \citep{zhao2023large}. It comprises 167,000 patient summaries paired with over 3.1 million patient-article relevance annotations and 293,000 patient-patient similarity links, derived from the citation graph of biomedical literature. The dataset supports two primary retrieval tasks: Patient-to-Article Retrieval (PAR), where the goal is to retrieve relevant scientific literature given a patient summary, and Patient-to-Patient Retrieval (PPR), which aims to identify clinically similar patient cases. Its scale, annotation quality, and diversity make it a valuable testbed for models that integrate semantic understanding and clinical relevance. PMC-Patients includes predefined training, validation, and test splits to facilitate reproducibility and standard evaluation across retrieval methods in the clinical domain.

\paragraph{Long Context Tasks}

\paragraph{i2b2: NER}  
The i2b2 datasets are a suite of de-identified clinical text corpora released through shared tasks organized by the Informatics for Integrating Biology and the Bedside (i2b2) initiative \citep{murphy2010serving}. These datasets span multiple years (e.g., 2006, 2010, 2012, 2014) and cover a range of clinical named entity recognition (NER) challenges, such as extracting problems, treatments, tests, and temporal expressions from patient narratives. Unlike synthetic or abstracted biomedical corpora, i2b2 datasets consist of real-world clinical notes, making them uniquely valuable for benchmarking models in practical clinical NLP settings. Due to their rich annotation schemas and long-form input structure, they are particularly well-suited for evaluating models on long-context NER and contextual disambiguation tasks. In our setup, we preserve the original document structure and avoid chunking, allowing Clinical ModernBERT to leverage its extended context capacity to model entity boundaries across sentences and paragraphs.

\section{Model Optimizations}
\label{mlm}

\subsection{MLM Accuracies and MLM Loss}

\begin{figure}[h!]
    \centering
    \includegraphics[width=0.7\linewidth]{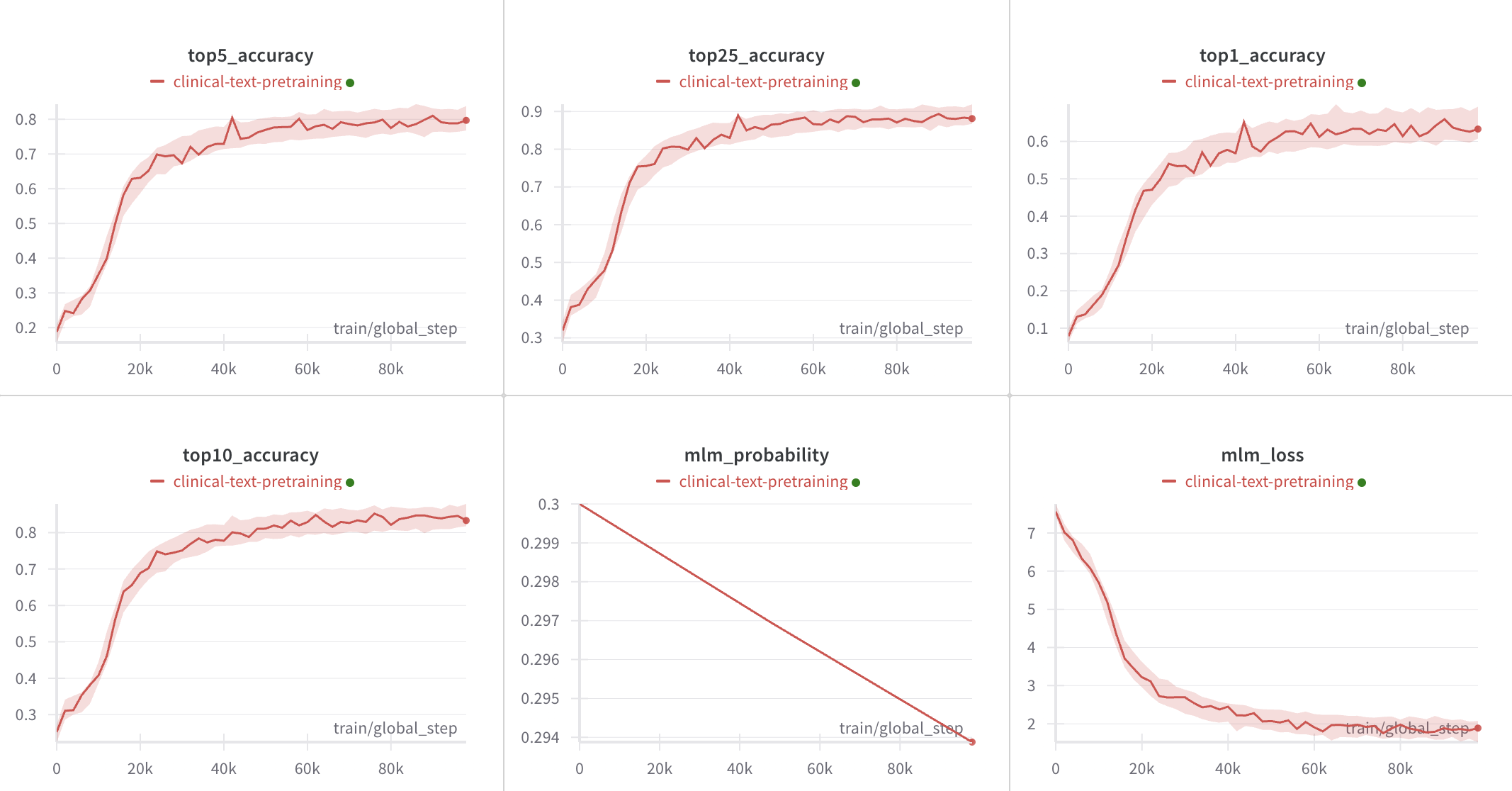}
    \includegraphics[width=0.8\linewidth]{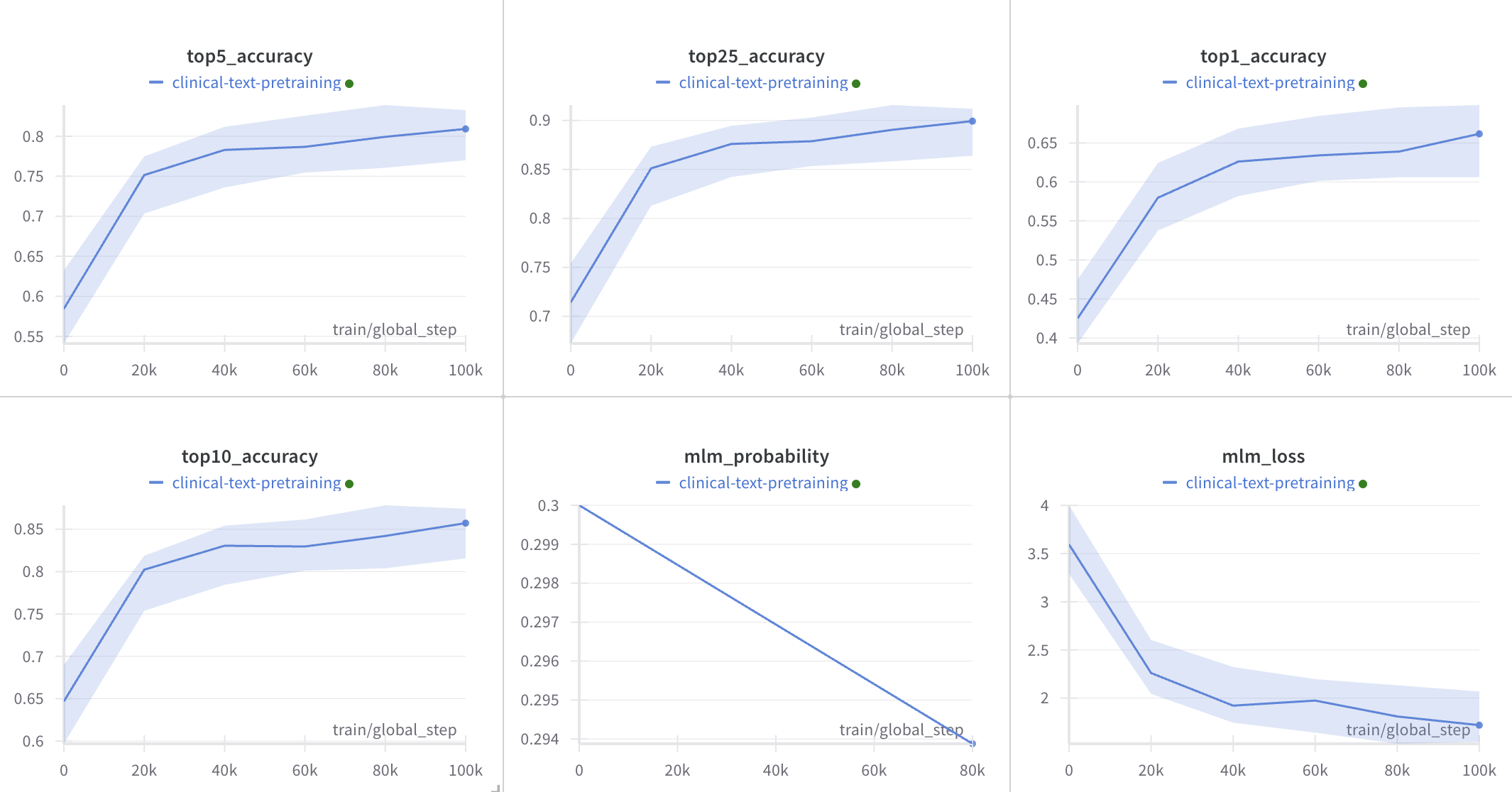}
    \includegraphics[width=0.8\linewidth]{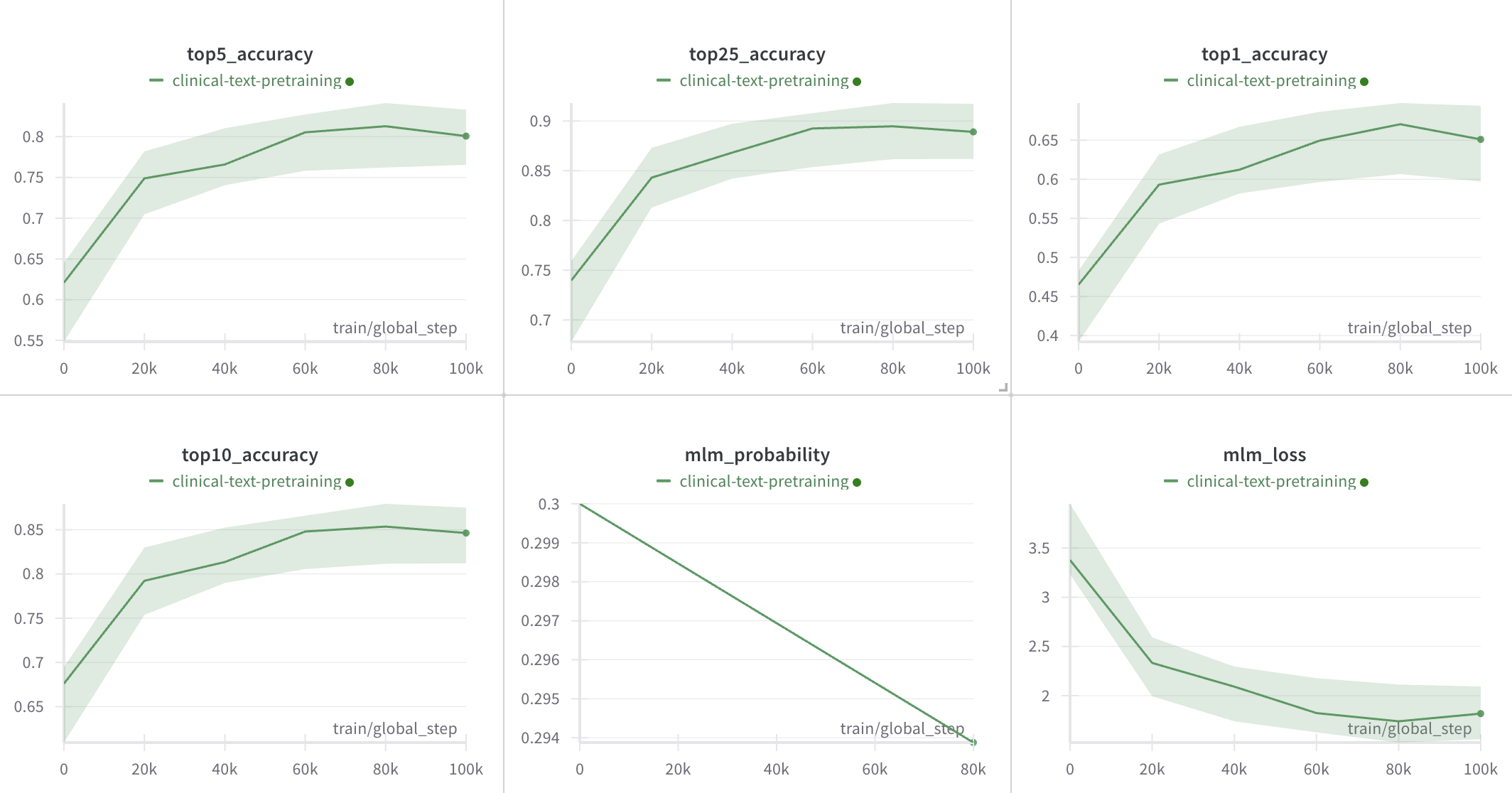}
    \caption{\textbf{Masked Language Modeling (MLM) Top-K Accuracies and Loss:} We report top-K accuracies for \(k = 1, 5, 10, 25\) alongside MLM loss across three pre-training runs initialized with different learning rates (top to bottom: \(3 \times 10^{-3}\), \(5 \times 10^{-4}\), \(1 \times 10^{-5}\)). Higher learning rates yielded more stable convergence and avoided shallow local minima, suggesting improved exploration of the loss landscape. As expected, larger learning rates also introduced noisier gradient updates, which aligns with standard intuitions in stochastic optimization.}
    \label{wandb}
\end{figure}

\newpage
\section{Pre-training Code and Model Weights}

\subsection{Source Code and Model Weights}

The full source code for pretraining, finetuning, and evaluation of Clinical ModernBERT is available at: 
\url{https://github.com/Simonlee711/Clinical_ModernBERT}.

Model weights for the pretrained Clinical ModernBERT checkpoint can be accessed via the Hugging Face Hub: 
\url{https://huggingface.co/Simonlee711/Clinical_ModernBERT}.

\end{document}